\documentclass[conference]{IEEEtran}
\IEEEoverridecommandlockouts
\usepackage{cite}
\usepackage{amsmath,amssymb,amsfonts}
\usepackage{algorithmic}
\usepackage{algorithm}
\usepackage{graphicx}
\usepackage{textcomp}
\usepackage{xcolor}
\usepackage{multirow}
\usepackage{multicol}
\usepackage[utf8]{inputenc}
\newcommand{\hw}[1]{\ensuremath{\mathtt{#1}}}

\usepackage{url}

\def\BibTeX{{\rm B\kern-.05em{\sc i\kern-.025em b}\kern-.08em
    T\kern-.1667em\lower.7ex\hbox{E}\kern-.125emX}}
\begin{document}

\title{Accelerating Framework of Transformer by Hardware Design and Model Compression Co-Optimization}

\author{\IEEEauthorblockN{
Panjie Qi\textsuperscript{1},
Edwin Hsing-Mean Sha\textsuperscript{1,*}, Qingfeng Zhuge\textsuperscript{1}, 
Hongwu Peng\textsuperscript{2},
Shaoyi Huang\textsuperscript{2},\\
Zhenglun Kong\textsuperscript{3},
Yuhong Song\textsuperscript{1}, 
Bingbing Li\textsuperscript{2} \thanks{* Corresponding Author}}

\IEEEauthorblockA{\textsuperscript{1}Department of Computer Science and Technology, East China Normal University, Shanghai, China}
\IEEEauthorblockA{\textsuperscript{2}Department of Computer Science and Engineering, University of Connecticut, Connecticut, USA}
\IEEEauthorblockA{\textsuperscript{3}Department of Electrical and Computer Engineering, Northeastern University, Boston  
\vspace{-0.15in}}
}



\maketitle

\begin{abstract}
State-of-the-art Transformer-based models, with gigantic parameters, are difficult to be accommodated on resource constrained embedded devices. Moreover, with the development of technology, more and more embedded devices are available to run a Transformer model. For a Transformer model with different constraints (tight or loose), it can be deployed onto devices with different computing power. However,  in previous work, designers did not choose the best device among multiple devices. Instead, they just used an existing device to deploy model, which was not necessarily the best fit and may lead to underutilization of resources. To address the deployment challenge of Transformer and the problem to select the  best device, we propose an algorithm$\leftrightarrows$hardware closed-loop acceleration framework. Given a dataset, a model, latency constraint $LC$ and accuracy constraint $AC$, our framework can provide a best device satisfying both constraints. In order to generate a compressed model with high sparsity ratio, we propose a novel pruning technique, hierarchical pruning (HP). We optimize the sparse matrix storage format for HP matrix to further reduce memory usage for FPGA implementation. We design a accelerator that takes advantage of HP to solve the problem of concurrent random access. Experiments on Transformer and TinyBert model show that our framework can find different devices for various $LC$ and $AC$, covering from low-end devices to high-end devices. Our HP can achieve higher sparsity ratio and is more flexible than other sparsity pattern. Our framework can achieve 37$\times$, 1.9$\times$, 1.7$\times$ speedup compared to CPU,GPU and FPGA,respectively.
\end{abstract}


\section{Introduction}

Recently, Transformer \cite{vaswani2017attention} has gained popularity and achieved record-breaking results on major natural language processing (NLP) tasks, including question answering, sentiment analysis and language inference \cite{wang2016inner,devlin2018bert,sukhbaatar2015end,rocktaschel2015reasoning}. Although state-of-the-art Transformer models offer great prediction accuracy,  they have a large number of parameters. For example, the $\text{BERT}_\text{LARGE}$ model has 340M parameters \cite{devlin2018bert} and the DistilBERT, a compact model, has 67M parameters \cite{sanh2019distilbert}. Moreover, with the ongoing democratization of machine learning \cite{garvey2018framework}, there are increasing needs to execute such giant models on embedded devices \cite{li2019edge,xu2018scaling}, e.g.,  field-programmable gate array (FPGA) or application-specific integrated circuit (ASIC). Using these devices as an acceleration platform for Transformer is challenging as they offer a limited on-chip memory and often possess limited off-chip bandwidth, both of which are critical for high performance. This restriction is particularly limiting for FPGA due to its extremely small on-chip memory, approximately 5MB for low-end FPGA (e.g., ZCU104) and 35MB for high-end FPGA (e.g., Alveo U200). Therefore, when Transformer comes to embedded devices, the primary challenge is in accommodating giant models onto these devices, along with the requirement of low inference latency.

\begin{table}[t]
\vspace{-0.1in}
\centering
\setlength{\tabcolsep}{3pt}
\caption{Comparisons of acceleration framework for Transformer models.
Our framework distinguishes from other works by  considering both AC and LC.}
\resizebox{\columnwidth}{!}{
\begin{tabular}{c|cc|c|c|c}
\hline
\multirow{2}[0]{*}{Methods} & \multicolumn{2}{c|}{Algorithm$\rightarrow$Hardware} & NAS   & Hardware & \multirow{2}[0]{*}{\textbf{Ours}}\\
& FTRANS\cite{li2020ftrans}  & \cite{guo2020accelerating} & HAT \cite{{wang2020hat}}   & A\textsuperscript{3} \cite{ham20203}  &  \\
\hline
AC \& LC   & $\times$   & $\times$  & $\times$    & $\times$ & \checkmark  \\
Hardware Type & $\times$     & $\times$     & \checkmark      & $\times$     & \checkmark  \\
Resource Uti. & \checkmark  & $\times$     & $\times$     & \checkmark   & \checkmark  \\
Compression & \checkmark & \checkmark      & $\times$     & $\times$    & \checkmark  \\
\hline
\end{tabular}}%
\label{tab:novel}%
\vspace{-0.25in}
\end{table}%

Three research trends have attracted enormous interests to improve the performance of Transformer, as Table \ref{tab:novel} shows. The first trend is hardware acceleration on ASIC, e.g., A\textsuperscript{3}~\cite{ham20203}, where researchers mainly focus on hardware acceleration. The second trend is algorithm optimization on CPU and GPU, such as neural architecture search (NAS) and model compression algorithms, e.g., block structured pruning~\cite{li2020_efficient}, lottery ticket hypothesis~\cite{chen2020lottery,prasanna-etal-2020-bert,wang2020hat}. 
The third trend is  the algorithm$\rightarrow$hardware sequential design flow \cite{li2020ftrans,guo2020accelerating}, which compress the model first and then implement compressed model to a existing device. This sequential design flow has no hardware performance feedback to software optimization. In this paper, we propose an algorithm$\leftrightarrows$hardware closed-loop framework, which can trade off between the sparsity ratio and hardware resources to achieve co-exploration of model compression and hardware acceleration. Our framework simultaneously considers hardware type, resource utilization, model compression, and $LC$ and $AC$. 

Moreover, with the development of technology, more and more hardware devices are available to run a Transformer model, such as various types of mobile device (e.g., Apple Bionic, Qualcomm Snapdragon, HiSilicon Kirin, Samsung Exynos, ...), FPGAs (e.g., ZCU102, VC707, Alveo U200, Versal, ...), ASICs and son on. These devices have different computing power and storage capacities, which are critical to the performance of Transformer. Moreover, for a Transformer model with different constraint requirements (tight or loose), it can be deployed onto devices with different computing power. However, in previous work, designers did not choose the best one among multiple devices. Instead, they just used an existing device to deploy the model, which was not necessarily the best fit and may lead to underutilization of resources. Therefore, with the surging of various types of devices and constraint requirements for models, it is becoming increasingly difficult for designers to select the best device for their application.

To address the deployment challenge of Transformer and the problem to select the best device, as the first attempt, we propose an algorithm$\leftrightarrows$hardware closed-loop framework, which can provide a best device under different constraints. Our framework makes a tradeoff between the sparsity ratio of model and hardware resources to achieve co-exploration to accelerate Transformer inference. We use FPGA to illustrate our design, and it can also be applied to other hardware devices, such as mobile devices, ASICs. 

The main contributions of this paper are: 
(1) \textbf{An Algorithm$\leftrightarrows$ hardware closed-loop framework.} We provide a co-exploration framework from constraints ($LC$,$AC$) to device. User can input some constraints, $LC$, $AC$, backbone model and dataset, our framework can output the best device to deploy this model meanwhile satisfying both constraints.
(2) \textbf{A Hardware-friendly Hierarchical Pruning (HP) Technique.} We propose HP, a novel sparsity pattern, which is a two-level pruning technique and takes a advantage of two existing pruning techniques, block structured pruning (BP) and vector-wise pruning (VW). HP is hardware-friendly and can achieve high sparsity ratio. (3) \textbf{A Sparse Matrix Storage Format Optimization.} We optimize a sparse weight format for our HP matrix on FPGA implementation. Our format can significantly reduce memory usage and perform better than commonly used formats. 
(4) \textbf{Sparsity-aware Accelerator.} We design a FPGA-based accelerator for HP  and  abstract a performance predictor to build a bridge between the software and hardware for efficient clock cycles and resource usage estimation.


\begin{figure*}[h]
    \vspace{-0.15in}
    \centering
    \setlength{\abovecaptionskip}{-0.1cm}
    \includegraphics[width=1.0\linewidth]{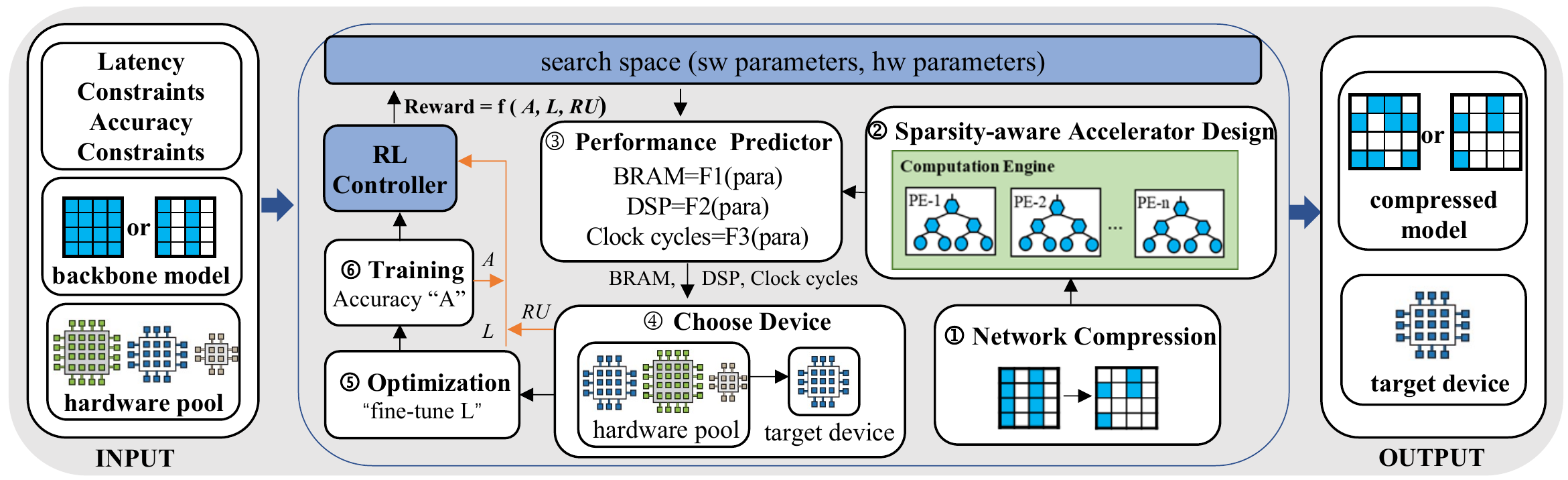}
    \caption{Overview of the proposed framework}
    \label{Figure:overview}
    \vspace{-0.2in}
\end{figure*}

\section{Related Work}
\textbf{Transformer.} 
Transformer has been highly optimized at the software level for CPU and GPU.
A research trend is to modify the architecture of Transformer to improve the performance on CPU and GPU \cite{wang2020hat,chen2018best}. These work exploit Neural Architecture Search (NAS) to search a best model architecture. However, the cost is usually high in the search process, since massive computations and neural network samples are required for an optimized network architecture. However,  little work has been published related to custom hardware acceleration for transformer-based model, particularly on FPGAs.  \cite{ham20203} has been proposed to  accelerate different parts of transformer model, attention and fully-connected layers, to achieve efficient processing on ASIC. \cite{li2020ftrans}  is the only currently published FPGA accelerator, which proposes a acceleration framework to enable compression on FPGA. This work sequentially first compress model and then deploy the compressed model on FPGA. This sequential design flow has no hardware performance feedback to software optimization and is not the optimal. In this paper, we trade off between the sparsity ratio and hardware resources to achieve co-exploration of model compression and hardware acceleration.


\textbf{Model Compression.} \cite{chen2020lottery,prasanna-etal-2020-bert} applied Lottery ticket hypothesis on model compression on BERT, based on an observation that a subnetwork of randomly-initialized network can replace the original network with the same performance. However, the non-structure pruning is not hardware-friendly. For hardware-friendly weight pruning, \cite{li2020efficient} proposes a hardware-friendly block structured pruning technique for transformer. However this technique will result in a significant accuracy loss when pruning ratio increases or block size is larger. \cite{ma2020pconv} proposes pattern pruning to make a better balance between accuracy and pruning ratio. But this pruning technique cannot directly apply to hardware due to parallelism limit. 

\textbf{Sparse Matrix Compression Formats.} A variety of sparse matrix representation formats have been proposed to compress the sparse matrix. Prior works take two major approaches to design such compression scheme. The first approach is to devise general compression formats, such as Compressed Sparse Row (CSR) \cite{2015CSR5}, Coordinate (COO) \cite{1995Templates}. They both record the row/column indices of each non-zero elements, which cause excessive memory usage.
The second approach is to leverage a certain known structure in a given type of sparse matrix. For example, the DIA format \cite{2009Pattern} is highly efficient in matrices where the non-zero elements are centrated along the diagonals of the matrix. The CSB format \cite{shi2020csb} is devised for the proposed CSB sparsity pattern. Though these compression schemes are specific to certain types of matrices, they are the most efficient in both computation and storage. In our work, in order to be the most efficient in storage, we optimize a sparse matrix compression scheme for our sparsity pattern HP.

\section{The Algorithm$\leftrightarrows$Hardware Closed-loop  Acceleration Framework}

To address the deployment challenge of Transformer and the problem to select the best device, as the first attempt, we propose an algorithm$\leftrightarrows$hardware closed-loop framework, which can provide a best device under different constraints. Our framework makes a tradeoff between the sparsity ratio and hardware resources to achieve co-exploration of model compression and hardware acceleration. Next, we use FPGA to illustrate our design, and it can also be applied to other devices, such as mobile devices, ASICs.

\subsection{Problem Definition and Overview}

In this paper, we aim to develop an algorithm$\leftrightarrows$hardware closed-loop acceleration framework to select the best device under different $AC$ and $LC$ for Transformer. We define the problem as follows: Given a specific data set $D$, a backbone model $bM$, a hardware pool $H$,  latency constraint $LC$, accuracy constraint $AC$, the objective is to determine: (i) $cM:$ a compressed model including sparsity of each layer; (ii) $tH:$ the target hardware device; such that the compressed model $cM$ can be deployed onto the target device $tH$ meanwhile satisfying both constraints $LC$ and $AC$.

Figure \ref{Figure:overview} shows the overview of our framework and Algorithm \ref{alg:framework} illustrates the whole process. Firstly, we design a pruning technique and conduct sparsity-aware accelerator design (components \textcircled{1} \textcircled{2}) and abstract a performance predictor to estimate hardware resource requirements (components \textcircled{3}). Then we use the RNN-based RL controller to guide  the  search  process: (i) the controller predicts a sample; (ii) the performance predictor roughly estimates  resource requirements of the sample  (components \textcircled{3}); (iii) select the target device from hardware pool to meet resource requirements (components \textcircled{4}); (iiii) fine tune the resource allocation exactly and optimize the latency under the target device constraint (components \textcircled{5}); (iiiii) train the model and get accuracy (components \textcircled{6}). At last, the controller is updated based the feedback (reward) from  \textcircled{4} \textcircled{5} \textcircled{6} and then predicts better samples. In the following text, we will introduce these components one-by-one.

\vspace{-0.1in}
\begin{algorithm}[htb] 
\small
\caption{Acceleration Framework.}
\label{alg:framework} 
\begin{algorithmic}[1] 
\REQUIRE 
$bM$: backbone model; $D$: a specific data set\\
~~~~ $LC$: latency constraint; $AC$: accuracy constraint \\
~~~~ $H$: a hardware pool
\ENSURE 
$cM:$ a compressed model\\
~~~~~ $tH:$ the target hardware
\FOR{each $i$ in range(1, $iterMax$)}
 \STATE RL controller predicts a sample $(sw,hw)$
\STATE Performance predictor to roughly predict clock cycles $E_{cycle}$, the number of block RAMs (BRAMs) $E_{bram}$ and DSPs $E_{dsp}$ based on the sample.
\STATE Choose device from $H$ based on $E_{cycle}$, $E_{bram}$, $E_{dsp}$.\\
\STATE Estimate the maximum latency $ML$
  \IF{find proper device and $ML < LC$}
  \STATE calculate the $RU$ and choose the best fit $tH$.
  \STATE fine tune resource allocation to get mini latency $L$
  \STATE $A,cM$ = Prune\_Train($bM$,$D$)
  \ELSE 
  \STATE assign negative values to $A,L,RU$
  \ENDIF
  \STATE Reward = $A$ + norm($L_{f}$) + $RU$
  \STATE Monte Carlo algorithm to update controller
\ENDFOR
\end{algorithmic}
\end{algorithm}
\vspace{-0.2in}


\subsection{Network Compression}
In order to accommodate Transformer models with enormous parameters onto the on-chip memory of FPGA, a pruning technique that can achieve a high sparsity ratio with a small accuracy loss and  hardware-friendly is necessary. In this paper, we propose HP, which is a two-level pruning technique. It combines the advantages of existing two pruning techniques, BP \cite{li2020_efficient} and VW \cite{cao2019efficient}. Firstly, to keep hardware-friendly, we adopt BP to prune model. However, it is coarse-grained and  can't achieve high sparsity ratio with a small accuracy loss. But how to achieve higher sparsity ratio? Next, based on BP, we adopt VW, a fine-grained pruning, to prune further. In this way, we can maintain hardware-friendly and achieve high sparsity.

\begin{table}[t]
\centering
\caption{Pruning techniques comparison among BP, VW and our HP. Our HP combines coarse-grained and fine-grained pruning and can achieve higher sparsity ratio than BP and VW.}
\setlength{\tabcolsep}{10pt}
\begin{tabular}{c|ccc}
\hline
& BP & VW & \textbf{HP(ours)} \\
\hline
Fine-grained & & \checkmark & \checkmark\\
Coarse-grained & \checkmark & &\checkmark \\
Flexibility & & & \checkmark\\
Hardware-friendly & \checkmark & \checkmark &\checkmark\\
High spar.\& acc. & & &\checkmark\\
\hline
\end{tabular}
\label{tab:pruning_table}
\vspace{-0.25in}
\end{table}

\begin{figure}[b]
    \centering
    \vspace{-0.15in}
    \setlength{\abovecaptionskip}{-0.1cm}
    \includegraphics[width=1.0\linewidth]{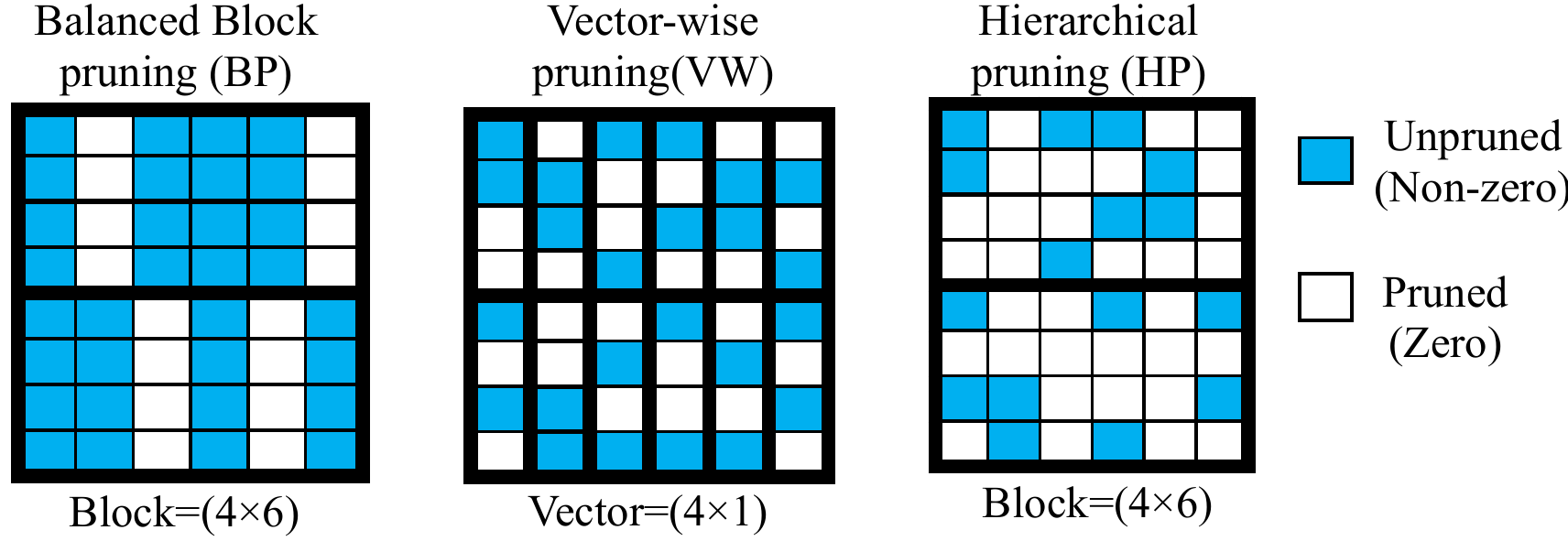}
    \caption{The Proposed Pruning Technique, Hierarchical Pruning (HP).}
    \label{figure:compression}
     \vspace{-0.15in}
\end{figure}

As Figure \ref{figure:compression} shows, our HP is a combination of BP and VW. First, we adopt BP as the first level pruning and we divide the weight matrix into blocks and prune some unimportant columns in each block. We regard this BP model as the backbone model of HP and denote its sparsity ratio by $S_{bm}$. The value of $S_{bm}$ determines the starting sparsity rate of HP weight, which can be adjusted flexibly. Then, based on the backbone model BP, we adopt VW as the second level to remove unimportant elements in each unpruned column of blocks.  To keep balanced, we remove the same number of elements in each column of blocks. Our HP combines coarse-grained (BP) and fine-grained pruning (VW) to achieve a higher sparsity ratio and ensure a small accuracy loss. We make a comparison among BP, VW and HP. As Table \ref{tab:pruning_table} shows, our HP combines the best of both BP and VW and is more flexible and effective than them. As for VW, it keeps all vectors (columns), which is unnecessary because some vectors are important and some are not. As for HP, we can first prune some unimportant columns, which can increase the sparsity ratio than VW to some extent. Moreover, we can also flexibly adjust the value of $S_{bm}$ to achieve different sparsity ranges and accuracy.

\subsection{Sparsity-aware Accelerator Design}\label{section:accelerator design}
In this section, first, we introduce the optimized sparse weight matrix storage format when implementing on FPGA. Then we introduce the accelerator design.

\noindent\textbf{The Storage Format Optimization.}
In sparse matrices, the number of non-zero elements (NZ) is much smaller than the number of zero elements. In order to avoid unnecessarily 1) storing zero elements and 2) performing computations on them, we need an efficient scheme to compress the sparse matrix. Various sparse matrix storage formats have been proposed, e.g., COO \cite{1995Templates}, CSR \cite{2015CSR5}, BCSR \cite{pinar1999improving}, Tile-Bitmap \cite{zachariadis2020accelerating}, MBR \cite{kannan2013efficient}. In our work, we use a bitmap format similar to MBR and optimize this format  based on our sparsity pattern HP to reduce memory usage further. 

\begin{figure}[b]
    \setlength{\abovecaptionskip}{-0.1cm}
    \vspace{-0.15in}
    \centering
    \includegraphics[width=1\linewidth]{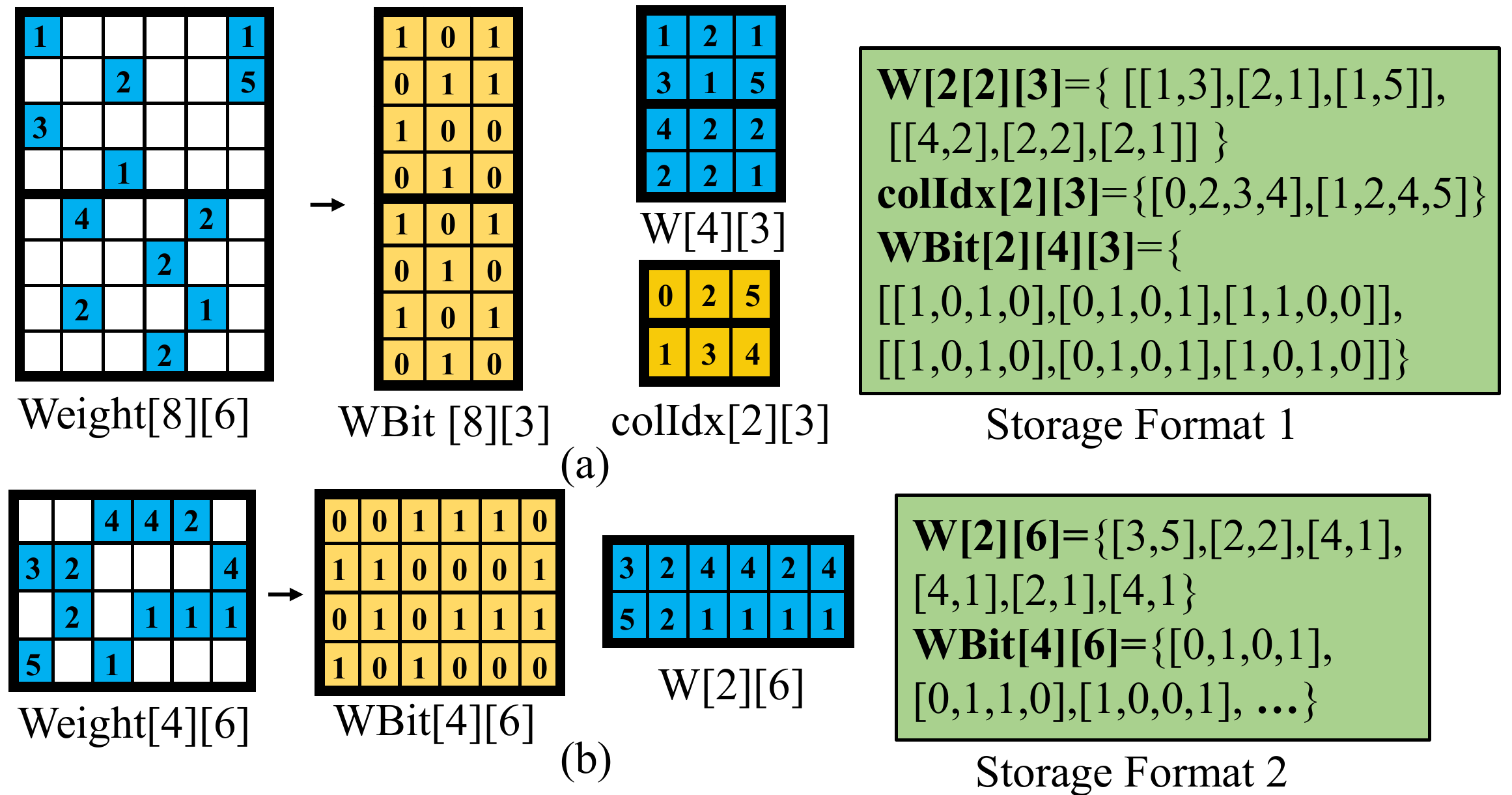}
    \caption{The Optimized Sparse Weight Matrix Storage Formats: WMark. (a) SF1 for $S_{bm}\neq0\%$. (b) SF2 for $S_{bm}=0\%$. $S_{bm}$ is the sparsity ratio of the backbone model (BP model).}
    \label{figure:storage_format}
    \vspace{-0.2in}
\end{figure}

Figure \ref{figure:storage_format} shows our format, WMark. We design two formats according to the sparsity ratio of backbone model $S_{bm}$. When $S_{bm}$ is not equal to 0\%, the weight is pruned by BP and VW and we use SF1. When $S_{bm}$ is equal to 0\%, the weight is only pruned by VW and we use SF2. There are three arrays in SF1: (i) the three-dimensional array \hw{W} records all non-zero elements (NZ). The first dimension record the number of blocks and the NZ in successive blocks are concatenated (column-major order) and stored continuously in the last two dimension; (ii) array \hw{colIdx} stores the indices of unpruned columns in each block; (iii) In order to track which elements are NZ in each unpruned column, we  use a bitmap array \hw{WBit}. If the a slot contains a NZ we set the it to "1", otherwise to "0". As for SF2, the \hw{colIdx} array is not needed and there are only two array. There are four arrays in MBR \cite{kannan2013efficient}: \hw{value, row\_Idx, col\_Idx, Bitmap}. The difference between our WMark and MBR \cite{kannan2013efficient} format is that: 1) \hw{row\_Idx} array is not needed. Because it is easy to calculate the row indices due to the balanced property of HP sparsity pattern. 2) The Bitmap array \hw{WBit} in our WMark only records the unpruned column not all columns, which can save memory storage. We set a $800\times800$ weight matrix with 50\% sparsity ratio and compare the memory usage of the five formats with ours. As Table \ref{tab:format_table} shows, our format performs better than all.

\begin{table}[t]
\vspace{-0.15in}
\centering
\setlength{\tabcolsep}{2pt}
\caption{Memory Usage Comparison among Sparse matrix formats.}
\resizebox{\columnwidth}{!}{
\begin{tabular}{ccccccc}
 \hline
& COO \cite{1995Templates} & CSR\cite{2015CSR5}   & BCSR \cite{pinar1999improving}  & Tile-Bitmap \cite{zachariadis2020accelerating}  & MBR \cite{kannan2013efficient} & \textbf{WMark(ours)} \\
 \hline
value & 1250 & 1250  & 2500 & 1250  & 1250  & \textbf{1250} \\
col\_Idx  & 3125 & 3125  & 312.5  & 312.5  & 312.5  & \textbf{312.5 } \\
row\_Idx & 3125  & 7.8   & 1.6   & 312.5  & 1.6   & \textbf{-} \\
Index & -  & -  & - & 351.6  & - & \textbf{-} \\
Bitmap & -  & -& - & 625 & 625  & \textbf{312.5 } \\
Total (Kb) & 7500  & 4382.8  & 2814.1  & 2851.6  & 2189.1  & \textbf{1875 } \\
\hline
\end{tabular}}%
\label{tab:format_table}%
\vspace{-0.2in}
\end{table}%

\begin{figure}
    \centering
    \setlength{\abovecaptionskip}{0cm}
    \includegraphics[width=1\linewidth]{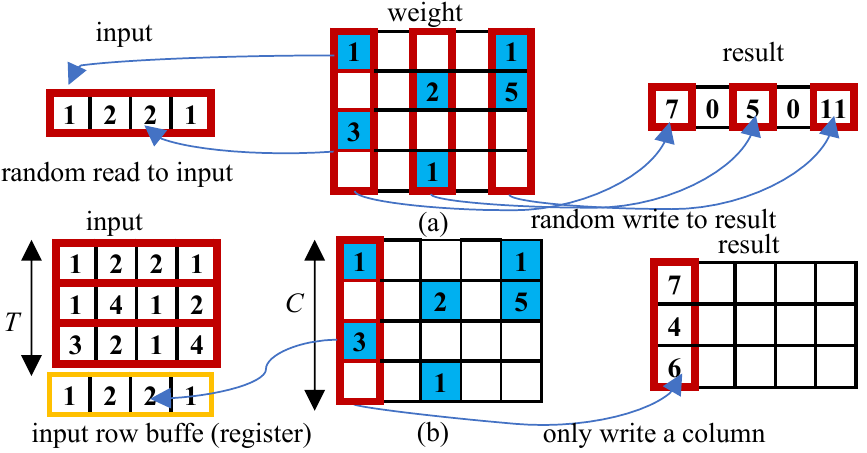}
    \caption{ The sparse MM parallelism scheme. (a) generalized VM and parallelism. (b) our MM design with HP. We exploit the multi-row of input matrix to avoid random write to result matrix.}
    \label{figure:parallelism_scheme}
\end{figure}

\noindent\textbf{Accelerator Design.}
Different from other FPGA accelerator design \cite{jiang2019accuracy,jiang2019achieving,jiang2019xfer, cao2019efficient}, we fit all weights on on-chip memory of FPGA and don't move data between on-chip and off-chip memory by  weight pruning and quantization. And to realize a low inference latency with parallel FPGA, there are multiple challenges to design an architecture that can exploit the benefits of HP. In previous work, \cite{cao2019efficient} and \cite{shi2020csb} implement accelerators with sparsity but they are designed for RNN model (matrix-vector multiplication, MV) and can't be applied to Transformer (matrix/vector-matrix multiplication, MM / VM). As Figure \ref{figure:parallelism_scheme} show, with generalized VM as in \cite{2018OuterSPACE}, there are two concurrent irregular memory accesses challenges, one for random read to input vector and the other for random write to result matrix, which can install the parallel execution. To solve these challenges, we change the memory access pattern. To avoid the random write to result matrix,  we multiply multiple rows of the input matrix by one column of the weight matrix, which can achieve sequential writing. To solve the challenge of random read to input, we assign a input matrix row buffer \hw{IRB} and use register to implement it which can be randomly accessed.

\begin{figure}
    \centering
    \setlength{\abovecaptionskip}{-0.1cm}
    \includegraphics[width=1\linewidth]{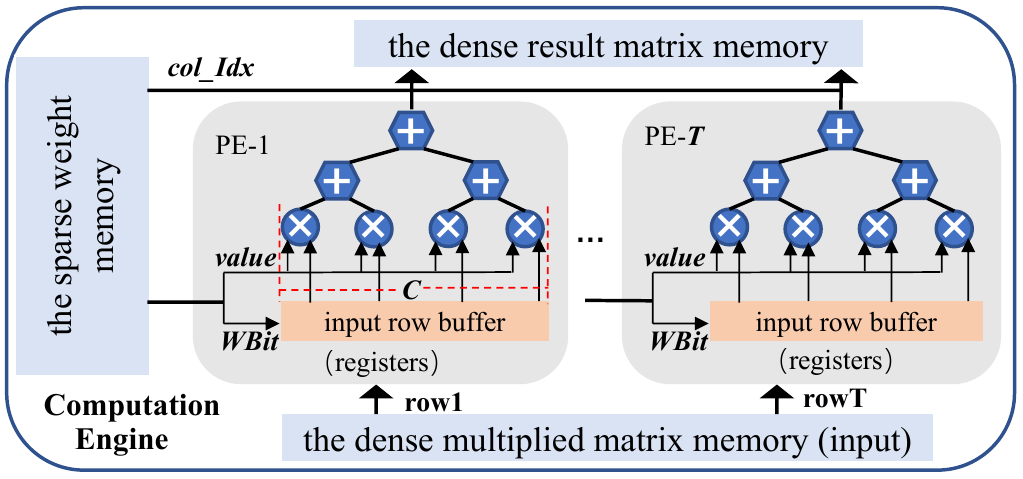}
    \caption{ Computation Engine.}
    \label{figure:on_chip_design}
    \vspace{-0.15in}
\end{figure}

Figure \ref{figure:on_chip_design} shows our computation engine. It consists of $T$ parallel processing elements (PEs) that compute dot products of distinct input matrix rows and one column of the weight matrix (one block) concurrently to exploit inter-row parallelism, while each PE is designed to exploit intra-row parallelism in a single dot product operation. Each PE contains a input matrix row buffer \hw{IRB} to buffer each row of the being multiplied input matrix and this buffer is implemented by register which  can be randomly accessed. This computation includes 5 steps: (1) The PE reads $C$ elements from the weight matrix memory and $C$ elements based on the \hw{WBit} array from the input row buffer \hw{IRB}. (2) $C$ multipliers operate simultaneously to obtain $C$ scalar products. (3) an adder tree sums $C$ scalar products to calculate the dot product. (4) PE reads \hw{col\_Idx} from the weight matrix. (5) The dot product result is written back to the result memory based on the \hw{col\_Idx}. PEs are fully pipelined so that one operation can be processed per clock cycle.

\subsection{Performance Predictor} \label{performance predictor}
We develop a FPGA performance predictor to roughly analyze resource requirements $E_{cycles}$, $E_{bram}$, $E_{dsp}$ based on software and hardware parameters predicted by RL controller.

\textcircled{1} We model the Block RAMs (on-chip SRAM units, called BRAMs) usage $E_{bram}$ using the formula in \cite{zhao2019performance}. According to on-chip buffer allocation, we can calculate the BRAMs for $i$-th buffer  $B_{i}=\left \lceil \frac{bits}{width} \right \rceil\ast \left \lceil \frac{elements}{depth} \right \rceil\ast factor$. Among them, $bits$ represents the quantization bits of weight and $elements$ represents the number of NZ of weight. The $width$ and $depth$ represent the configuration of BRAM. Then we can get the total BRAMs by adding up all buffers $E_{bram} = \sum B_{i}$.

\textcircled{2}The DSP usage $E_{dsp}$ is related to multiply-accumulate and data type. According to the computation engine in Figure \ref{figure:on_chip_design}, it can execute $C \times T$  MAC operations in parallel. For the 16-bit fixed point, it requires $2\times C \times T$ DSPs, where each multiplication and add operation requires 1 DSP. For 32-bit floating point, it requires $5\times C\times T$ DSPs, where 5 is the sum of 3 DSPs for one multiplication and 2 DSPs for one add operation \cite{TP-toolbox-web}. Suppose that the total number of layers are $n$  and the PEs size of $i$-th layer is $C_i \times T_i$, then the total DSP is :
$E_{dsp} = \sum_{i=1}^{n}5\times C_{i} \times T_{i}$.

\textcircled{3}The clock cycles $E_{cycles}$ are related to the size of PEs. After implementing PEs in Vivado HLS, we try to make the pipeline interval become 1, indicating that PEs can output one result in 1 clock cycles. Therefore, clock cycles of one layer equal the number of times that PEs is invoked. The sparse matrix multiplication of $K \times M$ and $M \times N$ with sparsity ration $s$ can support $K \times M \times N \times (1-s)$ MAC. With the PEs size $C \times T$, we can calculate the clock cycles: $l =\frac{K\times M\times N\times (1-s))}{T \times C}$. Therefore, for $n$ layers in total, the total clock cycles is: $E_{clock} = \sum_{i=1}^{n}l_{i}$.

\subsection{Choose Device}
Next, we introduce how to choose the best device from hardware pool based on $E_{cycles}$, $E_{bram}$, $E_{dsp}$. Figure \ref{fig:choose_device} show the process of selecting a best device. Table \ref{table:hardware_pool} show our hardware pool. The process is as follows: (1) First, we sort devices in hardware pool according to the number of BRAMs provided by each device. (2) we perform binary search to find the device whose BRAMs are large than $E_{bram}$. That might be more than one device thus we use a set $F$ to denote the alternative devices. (3) we calculate the latency $L_{i}$ for device $i$ in set $F$ based on the formula $L_{i}= E_{cycles} / freq_{i}$, where $freq_{i}$ is the running frequency of device $i$ and meanwhile we also compute the resource utilization $RU_{i}$ for each device. (4) we choose the device whose $L_{i}$ is small than $LC$. Specifically, When there are more than two device to choose from, we choose the device with largest $RU_{i}$, meaning that we can select the device with lower price and higher resource utilization.

\begin{table}
\setlength{\abovecaptionskip}{0.1cm}
\vspace{-0.1in}
\centering
\caption{Hardware Pool}
\begin{tabular}{c|c|c|c|c}
\hline
Devices & BRAM & DSP & LUT & FF \\
\hline
Alveo U200 & 4320  & 6840  & 21182240 & 2364480 \\
VC709 & 2940  & 3600  & 433200 & 866400\\
VC707 & 2060  & 2800  & 303600 & 607200\\
ZCU102 & 1824  & 2520  & 274080 & 548160 \\
ZCU104 & 624   & 1728  & 230400 & 460800\\ 
\hline
\end{tabular}%
\vspace{-0.1in}
\label{table:hardware_pool}
\end{table}%

\begin{figure}
    \centering
    \setlength{\abovecaptionskip}{-0.1cm}
    \includegraphics[width=1\linewidth]{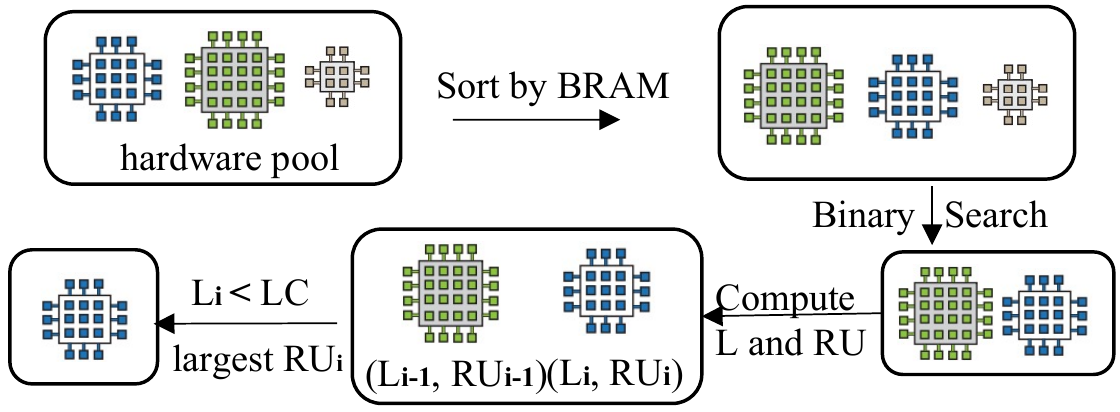}
    \caption{Choose target device}
    \label{fig:choose_device}
    \vspace{-0.2in}
\end{figure}

\subsection{Optimization}
The optimization step is to exactly fine tune the resource allocation sheme under the resource constraint of the target device to achieve the least clock cycles and fill up the gap between the actual and estimated value. In this step, we specifically consider the parallelism of the Dot-Attention layer, which defaults to 1 in the performance predictor step.  The target FPGA offers a certain number of resource including DSP slices (ALUs) and BRAMs. And our goal is to leverage these resources to exploiting the compute capabilities of the FPGA and achieving reasonably high performance. This problem can be expressed as the following  optimization objective:


\vspace{-0.2in}
\begin{equation*}
\small
\label{optimization}
\begin{split}
&\min \,\,\, exeCyc=g(T,C,h)=\sum_{i=0}^{n} f(T_{i},C_{i}) + \frac{nHead}{h}*Cyc_{h} \\
&s.t.\quad  \left\{\begin{array}{lc}
0\leq \sum_{i=0}^{n} C_{i} \times T_{i} + h*R_{h}\leq R_{total}\\
\end{array}\right.
\end{split}
\end{equation*}

\noindent Here, the parameters $T_{i}, C_{i}$ are the PE size of $i$-th layer and $h$ is the parallelism of Dot-Attention layers. The $Cyc_{h}$ and $R_{h}$  are the clock cycles and computation resource  needed by one Dot-Attention layer. $R_{total}$ is the avaliable computation resource of the target device. $nHead$ is the mumber of heads of Transformer-based models. Algorithm \ref{alg:optimization} illustrates our fine-tuned resource allocation scheme and solves the optimization objective. The algorithm takes in as input the Transformer model architecture $A$ and the target device constraints $F$. And it finally outputs the parallelism factor $C,T,h$ and the least latency $exeCycle$. 

\vspace{-0.05in}
\begin{algorithm}[htb] 
\small
\caption{ Fine-tuned Resource Allocation Scheme}
\label{alg:optimization} 
\begin{algorithmic}[1] 
\REQUIRE 
$F$: the target device constraints\\
~~~~ $A$: the Transformer model architecture\\
~~~~ $s$: the sparsity ratio for all layers
\ENSURE 
$exeCyc$: the optimized cycles\\
~~~~~~ $C,T,h$: parallelism factor
\STATE Initialize the $eec$ $\verb|//|$ execution\_cycles
\STATE Set avaliable computation resource: $R_{total}$  $\verb|//|$total DSPs
\STATE Compute the computation complexity of $i$-th layer: $Com_{i}= MAC\_of\_layer_i \times s_{i}$ and the total computation complexity of all layers: $Com_{total}=\sum_{i=0}^{n} Com_{i}$ 
\FOR{each $i$ in range(0, $A.numHead$)}
 \STATE $tempR = R_{total}-i*R_{h}$ 
 \FOR{each $j$ in range(0, $n$)}
    \STATE $R_{j}=\frac{Com_{j}}{Com_{total}} \times tempR$;
    \STATE adjust the parallelism factor ($C_{j} \times T_{j}$) based on $R_{j}$
 \ENDFOR
 \STATE calculate cycles = $g(C,T,h)$
 \IF{cycles $<$ $eec$}
  \STATE $eec$=cycles
  \STATE record $C,T,i$
\ENDIF
\ENDFOR
\end{algorithmic}
\end{algorithm}
\vspace{-0.15in}

\subsection{Reinforcement Learning (RL)}
In our design, the search space is very big, therefore we exploit the RL to carry out guided search. The RL controller is implemented based on an RNN \cite{zoph2016neural}.  In each episode, the controller first predicts a sample, and gets its $Reward$ based on the evaluation results from the environment (components \textcircled{3} \textcircled{4} \textcircled{5} \textcircled{6} in Figure \ref{Figure:overview}). Then, we employ the Monte Carlo policy gradient algorithm \cite{williams1992simple,yang2020co} to update the controller:
\[\tiny \nabla J\left ( \theta_{c} \right )= \frac{1}{m}\sum_{k=1}^{m}\sum_{t=1}^{T} \gamma^{T-t} \nabla_{\theta } log \left (a_{t}|a_{(t-1):1,} \theta_{c} \right )\left ( R_{b}-b \right )\]
where $m$ is the batch size and $T$ is the number of steps in each episode. The exponential factor $\gamma$ are used to adjust the reward at every step and the baseline $b$ is the average exponential moving of rewards.

\begin{figure}
    \centering
    \vspace{-0.15in}
    \setlength{\abovecaptionskip}{-0.1cm}
    \includegraphics[width=1\linewidth]{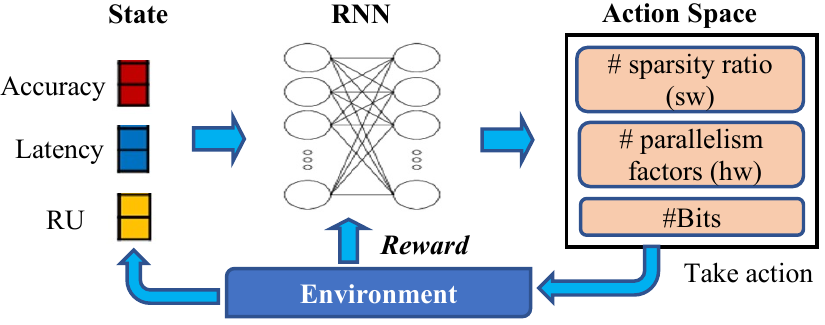}
    \caption{Reinforcement Learning. We integrate the software  parameters (\#sparsity ratio) and hardware parameters (\#parallelism factors) into the action space to realize a co-exploration of sparsity ratio and hardware resource. The environment is made up of  components \textcircled{3} \textcircled{4} \textcircled{5} \textcircled{6} in Figure \ref{Figure:overview}.}
    \label{figure:RL}
    \vspace{-0.15in}
\end{figure}

Our framework specifically takes hardware performance ($L, RU$) into consideration rather than just model accuracy $A$. As Figure \ref{figure:RL} shows, we integrate the software parameters (\# sparsity ratio) and accelerator design parameters (\# parallelism factors) into the action space of controller to realize a co-exploration of sparsity ratio and hardware resources. Therefore, we employ a reward function to calculate  $Reward$, which takes the accuracy $A$, latency $L$, the resource utilization $RU$ and latency constraint $LC$  to calculate reward. The function is defined as follows:
\[Reward=\left\{
\begin{matrix} 
A+\frac{LC-L}{LC}+RU & L<LC,A>AC \\ 
-pen_{A} / -pen_{L}  & otherwise \\
\end{matrix}\right.\]
In the above function, there are two cases. First, if $L<LC $ and $A>AC$, it indicates that the performance of the sample can satisfy the constraints, we sum up the reward of hardware performance and accuracy. Otherwise, in any other case, it indicates that the sample can't satisfy constraints and we directly return negative values to the controller, which can save the search time. Note that we return different negative reward to guide the search. We return $-pen_{A}$ for $L<LC \& A<AC$ and return $-pen_{L}$ for $L>LC \& A>AC$.

\section{Experiments}
\subsection{Experimental Settings}

\textbf{Baseline Models and Datasets.} We test our method on Transformer model using WikiText-2 dataset \cite{merity2016pointer} and on TinyBERT model using GLUE benchmark \cite{wang2018glue}. For Transformer model, there are 2 encoder and 1 decoder layers (the hidden size is 800, the feed-forward size is 200 and the head number is 4). And we use the accuracy of word prediction as our evaluation metrics. For TinyBERT, there are 4 encoder layers and 1 pooler layer and 1 classifier layer. 

\textbf{Evaluation Platforms.} We conduct the reinforcement learning framework with the training of Transformer model on an 8× NVIDIA Quadro RTX 6000 GPU server (24 GB GPU
memory). Experiments environment are performed on Python 3.6.10, GCC 7.3.0, PyTorch 1.4.0, and CUDA 10.1. The hardware accelerator design is implemented with Vivado HLS, which is the commonly used high level synthesis tool. This tool enables implementing the accelerator with C languages and exports the RTL as a Vivado's IP core. The C code of our accelerator is parallelized by adding HLS-defined pragma. Pre-synthesis resource report are used for performance estimation.

\subsection{Experimental Results}

\subsubsection{Pruning Strategy}
We set the sparsity ratio of backbone model $S_{bm}$ to different values for different models. For TinyBert model, its model size is relatively small and is sensitive to pruning. Therefore in order to maintain high accuracy,  we set $S_{bm}$ to $0\%$. For Transformer model, through experiments we set $S_{bm}$ to $50\%$, which can ensure high sparsity ratio and maintain acceptable accuracy loss. As Table \ref{tab:pruning_strategy} shows, Transformer model pruned by HP can reduce model size by $90\%$ with $2.37\%$ accuracy loss. And the TinyBert model can achieve $0.7\%$ and  $2.23\%$ accuracy loss for MRPC task and SST-2 task.

\textbf{Accuracy.} To evaluate  benefit of our HP, we compare it with BP \cite{li2020_efficient}, VW \cite{cao2019efficient}, block-wise pruning (BW) \cite{narang2017block}, and irregular pruning on Transformer model. The block size of BW and BP is $12 \times 12$, $10 \times 800$, respectively. And the vector size for VW is $10 \times 1$. As Figure \ref{fig:transformer_HP} shows, the HP, VW and the irregular pruning can achieve the same model accuracy when the sparsity is smaller than 70\%. The HP can achieve better accuracy than irregular pruning at around 82\% sparsity. When the sparsity is larger than 92\% (the intersection of HP and irregular), HP performs worse than irregular due to large sparsity of the backbone model. VW can only achieve the limited 90\% sparsity when the vector size is $10\times 1$ and our HP can achieve 99\% sparsity. These experimental results demonstrate that HP has almost the same effectiveness as irregular sparsity and outperforms BW, BP and VW sparsity in terms of achievable accuracy or sparsity during pruning.

\begin{table}[t]
\centering
\vspace{-0.15in}
\setlength{\tabcolsep}{1pt}
\setlength{\abovecaptionskip}{0.1cm}
\caption{Pruning Strategy}
\resizebox{\columnwidth}{!}{
\begin{tabular}{c|cc|cc|cc}
\hline
\multirow{2}{*}{} & \multicolumn{2}{c|}{Transformer} & \multicolumn{2}{c|}{TinyBert(MRPC)} & \multicolumn{2}{c}{TinyBert(SST-2)} \\
 & Base & HP($S_{bm}$=50\%) & Base & HP($S_{bm}$=0\%) & Base & HP($S_{bm}$=0\%) \\
\hline
model size & 52M   & 6M    & 14.5M & 10.9M & 14.5M & 8.6M \\
sparsity & 0.00\% & 89.85\% & 0.00\% & 25.0\% & 0.00\% & 41.00\% \\
accuracy & 98.50\% & 96.13\% & 86.45\% & 85.75\% & 92.60\% & 90.37\% \\
\hline
\end{tabular}
}
\label{tab:pruning_strategy}%
\vspace{-0.2in}
\end{table}%

\begin{figure}[b]
    \centering
    \vspace{-0.15in}
    \includegraphics[width=0.9\columnwidth]{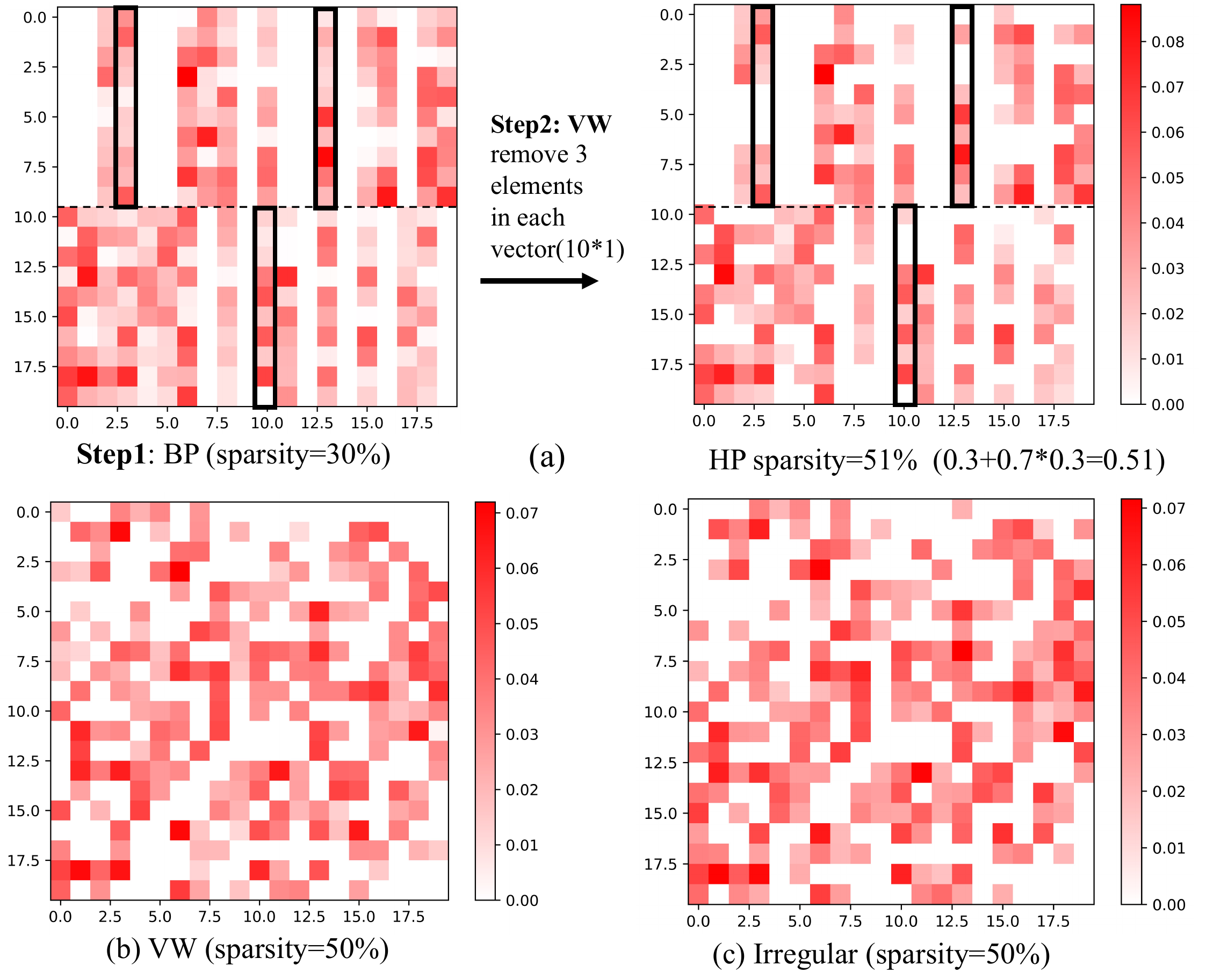}
    \caption{Weight heat map visualization after pruning with (a) HP (b) VW (c) irregular pruning. These weight heat maps are $20 \times 20$ sub-matrices of the whole $200 \times 800$ matrix of the feed forward layer2 in the encoder.}
    \label{fig:heatmap}
    \vspace{-0.25in}
\end{figure}

\textbf{Visualization.} We visualize the weight matrices after HP, VW and irregular pruning on Transformer model. Figure \ref{fig:heatmap} visualizes the three sparse weight matrices of a $20\times20$ sub-matrix which is randomly selected from the whole $200\times800$ weight matrix. Pink grids indicate non-zero parameters and the pink level indicates the magnitude of the absolute value. Figure \ref{fig:heatmap}(a) shows the two steps of HP. In our HP matrices, there are two blocks (the top and bottom of the dashed line) and each vector (column) of $10 \times 1$ in blocks has 7 NZ. We can see that the heat map of HP weight  can prune some unimportant columns and maintain most important weights as irregular pruning.  Although irregular sparsity retains some weights in a column while our HP removes the whole column, these weights are relatively small (this can be seen from the pink level in Figure \ref{fig:heatmap}) and the removal has no significant impact on accuracy. Instead, most of the important weights can be retained by our HP to ensure accuracy.

\begin{figure}
    \centering
    \setlength{\abovecaptionskip}{-0.1cm}
    \includegraphics[width=0.9\columnwidth]{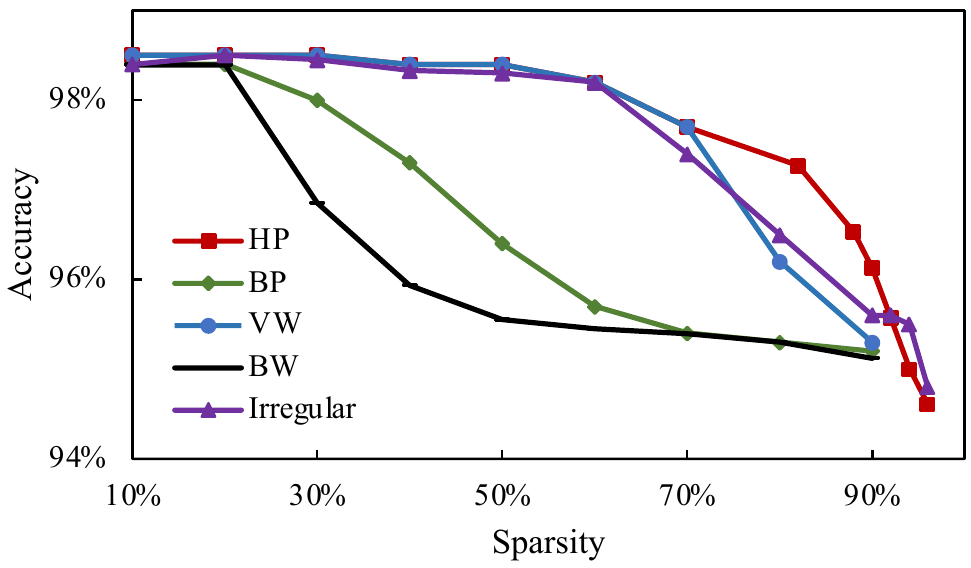}
    \caption{Accuracy comparison of Transformer model on WikiText-2 dataset with various pruning patterns.}
    \label{fig:transformer_HP}
    \vspace{-0.15in}
\end{figure}

\subsubsection{ Overhead Comparison of Sparse Weight Format}
\begin{figure}[b]
    \vspace{-0.15in}
    \centering
    \setlength{\abovecaptionskip}{-0.1cm}
    \includegraphics[width=0.95\linewidth]{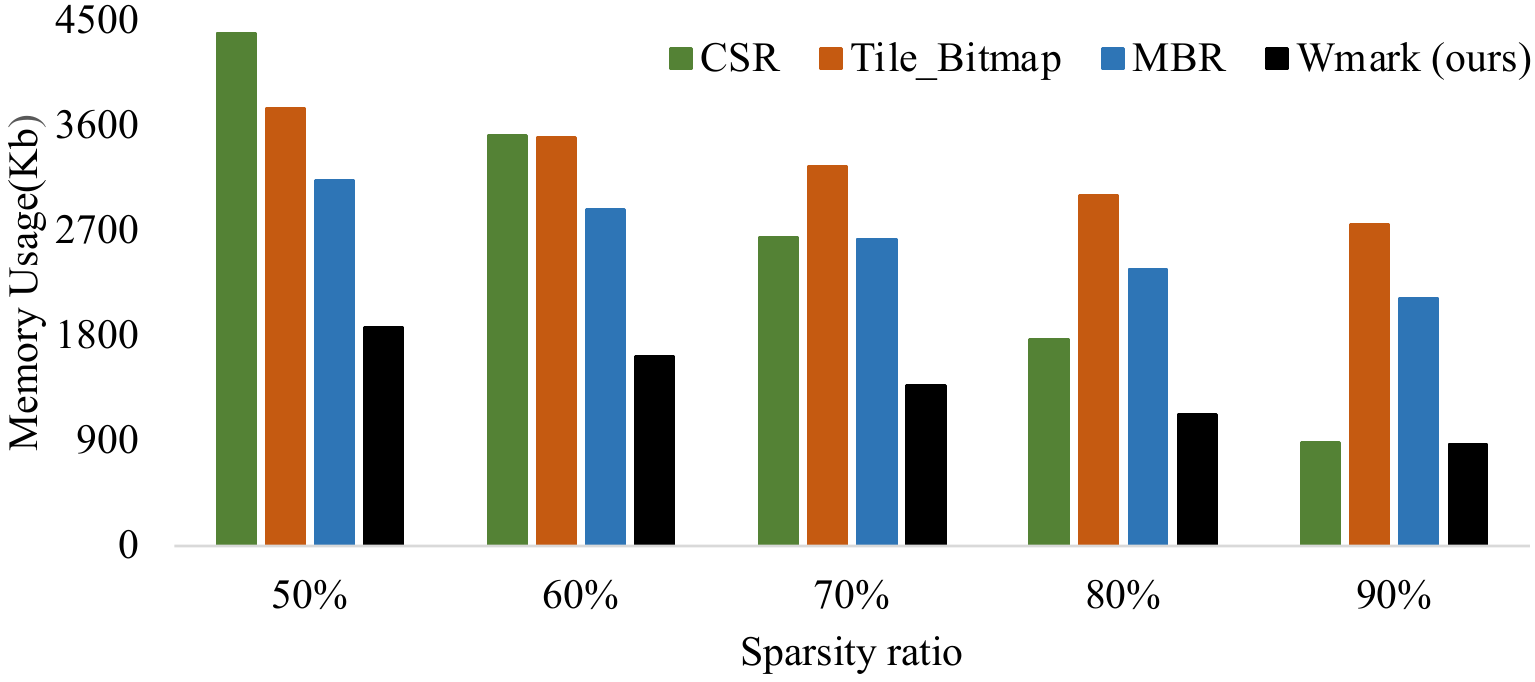}
    \caption{Overhead comparison of sparse weight matrix formats among CSR, Tile-Bitmap, MBR and our WMark.}
    \label{figure:storage_format_comparison}
     \vspace{-0.15in}
\end{figure}

We compare overhead among CSR \cite{2015CSR5}, Tile-Bipmap \cite{zachariadis2020accelerating}, MBR \cite{kannan2013efficient} and our WMark format. We use the memory usage as the metric. Figure \ref{figure:storage_format_comparison} shows the results, it is clear that our optimized format WMark needs the least memory than all of them. And the WMark can achieve $1.5 \times -2.5\times$  reduction in memory usage than MBR \cite{kannan2013efficient}. The reason is that our WMark has the balanced property and we don't need the \hw{row\_start} array to calculate the start index of each row. Besides, our \hw{WBit} array only mask the non-zero columns not all columns. Therefore, our WMark can use less memory than MBR \cite{kannan2013efficient}.


\begin{table*}
\vspace{-0.1in}
\centering
\setlength{\tabcolsep}{4pt}
\setlength{\abovecaptionskip}{0.1cm}
\caption{Validation On FPGA}
\resizebox{\textwidth}{!}
{\begin{tabular}{cccccccccc}
\hline
\textbf{Models} & \textbf{(LC,AC)}& \textbf{sparsity} & \textbf{accuracy} & \textbf{est. latency} & \textbf{BRMA / Uti} & \textbf{DSP / Uti} & \textbf{LUT / Uti} & \textbf{FF / Uti} & \textbf{Target device} \\
\hline
\multirow{2}[2]{*}{Transformer} 
& (40ms, 92\%) & 92.00\% & 94.45\% & 35.70ms & 2492 / 85\% & 1644 / 46\% & 303879 /70\% & 268065 / 30\% & VC709 \\
& (20ms, 96\%) & 86.42\% & 96.84\% & 18.10ms & 3311 / 77\% & 5040 / 74\% & 908833 / 77\% & 1102880 / 47\% & Alveo U200 \\
\hline
{\multirow{4}{*}{\shortstack{TinyBert\\(MRPC)}}} & (180ms, 85\%) & 0\%  & 86.45\% & 175ms & 1602 / 87\% & 1027 / 40\% & 262248 / 95\% & 131542 / 23\% & ZCU102 \\
& (45ms, 85\%) & 0\%   & 86.45\% & 42.1ms & 2194 / 74\% & 1928 / 53\% & 417058 / 96\% & 254178 / 29\% & VC709 \\
& (18ms, 85\%) & 0\%   & 86.45\% & 15.8ms & 4204 / 97\% & 4145 / 60\% & 936545  / 79\% & 504293 / 21\% & Alveo U200 \\
& (50ms, 80\%) & 27.00\% & 84.95\% & 47.33ms & 1530  / 83\% & 991 / 39\% &  254543  / 92\% & 158817  / 28\% & ZCU102 \\
\hline
{\multirow{2}[2]{*}{\shortstack{TinyBert\\(SST-2)}}} & (45ms, 90\%) & 25.00\%  & 90.83\% & 40ms  & 1674 / 91\% & 1056 / 42\% & 264443 / 96\% & 189035 / 34\% & ZCU102 \\
& (30ms, 90\%) & 41.00\%  & 90.37\% & 25.1ms & 2504 / 85\% & 2028 / 56\% & 316028 / 73\% & 235177 / 27\% & VC709 \\
\hline
\end{tabular}}%
\label{tab:validation_on_FPGA}%
\vspace{-0.2in}
\end{table*}%

\subsubsection{Validation On FPGA}
We use FPGA devices to validate our approach, Table \ref{tab:validation_on_FPGA} show our results.
Our approach can find different devices under different sets of $LC$ and $AC$. For Transformer, we set two sets of constraints to choose device: (40ms,92$\%$) for loose constraints and (20ms,96$\%$) for tight constraints. For (40ms,92$\%$), its latency and accuracy restrictions are loose, so it is possible to achieve a higher sparsity ratio and deploy to a mid-end FPGA VC709. For (20ms,96$\%$), its latency and accuracy restrictions are relatively tight, therefore the sparsity ratio is relatively small to ensure accuracy and it will be deployed to a device with strong computing power, Alveo U200, to achieve very low latency.

For MRPC task of TinyBERT model, first we set up three sets of constraints which have the same $AC$ but different $LC$. The experimental results show that constraints with smaller $LC$ can be deployed on FPGAs with greater computing power, such as (180ms, 85\%) to ZCU102, (45ms, 85\%) to VC709. Then we set constraint with lower $AC$ (50ms, 80\%), the searched result of sparsity ratio is 25\% and the target device is ZCU102. This constraint can also be mapped to low-end FPGA (ZCU102), the same device as (180ms, 85\%), due to compression and can achieve $3.7 \times$ latency reduction. Therefore, the same device can satisfy different sets of constraints by compression and can be applied to different application scenarios. For SST-2 task of TinyBERT model, it shows similar experimental results as Transformer and MRPC task.


\subsubsection{Cross-platform Comparison}
The research on Transformer models mainly focus on at the software level for CPU and GPU, such as Trans \cite{vaswani2017attention}, Evolved Transformer \cite{so2019evolved}, and HAT \cite{wang2020hat}, but little work has been published related to custom hardware acceleration on FPGAs, in addition to FTRANS \cite{li2020ftrans}. We compare the efficiency of ours with these work.
Since these work exploit different models and data set, in order to provide a fair comparison, we use the floating-point operations per second (FLOPS) as the metric.
As Table \ref{tab:platform_comparison} shows, our FPGA implementation can achieve $31\times, 37\times, 27\times$ speedup compared to Trans \cite{vaswani2017attention}, Evolved Transformer \cite{so2019evolved} and HAT \cite{wang2020hat} on CPU respectively. And it can achieve $1.9\times$ and $1.7\times$ speedup compared to HAT \cite{wang2020hat} on GPU and FTRANS \cite{li2020ftrans} on FPGA. 


\begin{table}[b]
\centering
\vspace{-0.2in}
\setlength{\tabcolsep}{3pt}
\setlength{\abovecaptionskip}{0.1cm}
\caption{Comparison among CPU,GPU and FPGA}
\resizebox{\columnwidth}{!}{
\begin{tabular}{c|ccc|c|c|cc}
\hline
 & \multicolumn{3}{c|}{CPU} & GPU & FPGA & \multicolumn{2}{c}{\textbf{Ours}} \\
 \multirow{2}{*}{} & \cite{vaswani2017attention} & \cite{so2019evolved} & \cite{wang2020hat} & \cite{wang2020hat} & \cite{li2020ftrans} & Trans(84\%) & Tinybert(0\%) \\
\hline
Operations(G) & 1.5   & 2.9   & 1.1   & 1.1  & 0.284 & 0.09  & 1.2 \\
latency(s) & 3.3   & 7.6   & 2.1   & 0.147 & 0.034 & 0.00645 & 0.0158 \\
FLOPS(G) & 0.45  & 0.38  & 0.52  & 7.48  & 8.35 & 14.14  & 75.94  \\ 
Impro. & base  & 0.84$\times$ & 1.16$\times$ & 16.62$\times$ &  18.5$\times$ & 31.42$\times$ & 168.75$\times$ \\
\hline
\end{tabular}
}%
\label{tab:platform_comparison}%
\vspace{-0.15in}
\end{table}%

\subsubsection{Search Space Exploration}\label{section:search space exploration} 
\begin{figure}
    \centering
    \setlength{\abovecaptionskip}{-0.1cm}
     \scalebox{1}
    {\includegraphics{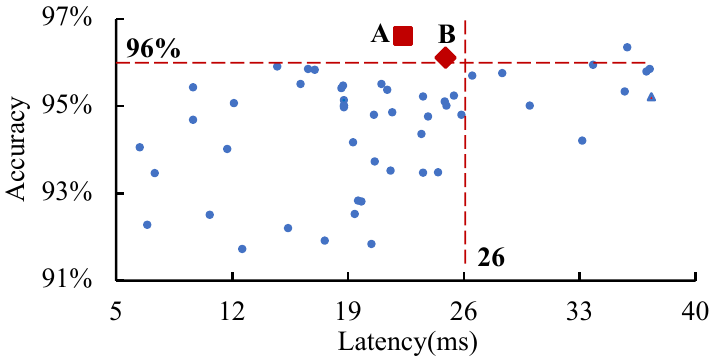}}
    \caption{ The RL Search Results for Transformer under constraint (26ms, 96$\%$). We select the device with the largest $RU$ from A and B. }
    \label{fig:search_20}
    \vspace{-0.2in}
\end{figure}

We collect the explored results from RL to form the search space exploration results of Transformer in Figure \ref{fig:search_20}. In this figure, the x-axis and y-axis stand for the latency and accuracy. We show the search result of constraint (26ms,96$\%$). We can see that these points are mainly concentrated in the vicinity of (26ms,96$\%$). This is due to the  guided search of RL, which makes the search samples closer to the solution. There are two points A and B satisfying the constraints in Figure \ref{fig:search_20}, it may represent there are two devices to choose from. In this case, we use the third metric, resource utilization, to select the best solution. The device with the largest resource utilization is the one that is more suitable, and at the same time it will be the cheaper one. Therefore, with our approach we can choose the best device.

\subsubsection{Ablation Study}\label{section:ablation_study}
\begin{figure}[!h]
\centering
\vspace{-0.1in}
\setlength{\abovecaptionskip}{-0.1cm}
\scalebox{1}
{\includegraphics{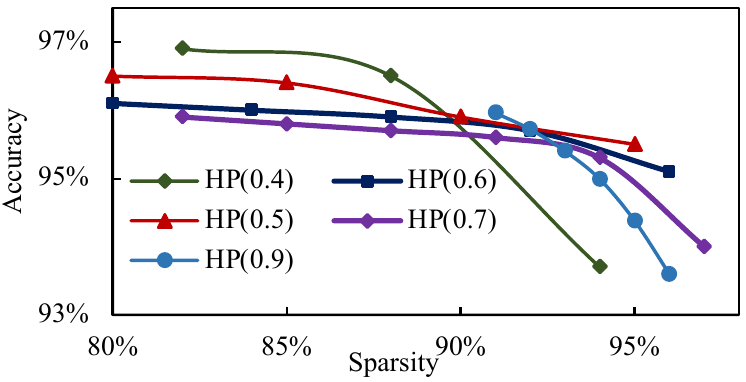}}
\caption{The accuracy comparison of HP with different backbone models.}
\label{fig:ablation_study}
\vspace{-0.15in}
\end{figure}

In this section, we investigate the influence of the sparsity of backbone model $S_{bm}$ in our HP. We set the row of block size as 10 in our experiment. The first feature of HP is that it can achieve different sparsity range when combined with different backbone models. For example, when $S_{bm}$ is equal to $40\%$, it can achieve sparsity range from $46\%$ to $94\%$. when $S_{bm}$ is equal to $80\%$, the sparsity range is from $82\%$ to $98\%$. VW has the limited sparsity of $90\%$  and it can't achieve sparsity larger than $90\%$.  For accuracy, as Figure \ref{fig:ablation_study} shows, under the same sparsity ratio, accuracy decreases with the increase of $S_{bm}$. For example, When sparsity is $88\%$, $40\%$ backbone model can achieve $96.4\%$, which is higher than $95.40\%$ and $95.20\%$ for $60\%$ and $70\%$  backbone model. So that tells us that we'd better make the value of $S_{bm}$ small in order to get a better accuracy.

\section{Conclusion}
In this paper, we propose an algorithm$\leftrightarrows$hardware closed-loop  acceleration framework to solve the challenge of efficient deployments and device selection problem. Our framework can achieve from constraints ($LC,AC$) to device. To achieve high sparsity ratio, we propose HP to reduce model size. To further reduce memory usage, we optimized the sparse matrix storage format based HP sparsity pattern. Experiments show that our framework can find different devices for various $LC$ and $AC$, covering from low-end devices to high-end devices.

\section*{Acknowledgments}
This work is partially supported by National Nature Science Foundation of China (NSFC) 61972154 and Shanghai Science and Technology Commission Project 20511101600. We sincerely thank Prof. Caiwen Ding at UConn and Prof. Weiwen Jiang at George Mason University for the intensive discussions and constructive suggestions.

\newpage
\bibliographystyle{IEEEtran}
\bibliography{ref}

\begin{thebibliography}{10}
\providecommand{\url}[1]{#1}
\csname url@samestyle\endcsname
\providecommand{\newblock}{\relax}
\providecommand{\bibinfo}[2]{#2}
\providecommand{\BIBentrySTDinterwordspacing}{\spaceskip=0pt\relax}
\providecommand{\BIBentryALTinterwordstretchfactor}{4}
\providecommand{\BIBentryALTinterwordspacing}{\spaceskip=\fontdimen2\font plus
\BIBentryALTinterwordstretchfactor\fontdimen3\font minus
  \fontdimen4\font\relax}
\providecommand{\BIBforeignlanguage}[2]{{%
\expandafter\ifx\csname l@#1\endcsname\relax
\typeout{** WARNING: IEEEtran.bst: No hyphenation pattern has been}%
\typeout{** loaded for the language `#1'. Using the pattern for}%
\typeout{** the default language instead.}%
\else
\language=\csname l@#1\endcsname
\fi
#2}}
\providecommand{\BIBdecl}{\relax}
\BIBdecl

\bibitem{vaswani2017attention}
A.~Vaswani, N.~Shazeer, N.~Parmar, J.~Uszkoreit, L.~Jones, A.~N. Gomez,
  {\L}.~Kaiser, and I.~Polosukhin, ``Attention is all you need,'' in
  \emph{Advances in neural information processing systems}, 2017, pp.
  5998--6008.

\bibitem{wang2016inner}
B.~Wang, K.~Liu, and J.~Zhao, ``Inner attention based recurrent neural networks
  for answer selection,'' in \emph{Proceedings of the 54th Annual Meeting of
  the Association for Computational Linguistics (Volume 1: Long Papers)}, 2016,
  pp. 1288--1297.

\bibitem{devlin2018bert}
J.~Devlin, M.-W. Chang, K.~Lee, and K.~Toutanova, ``Bert: Pre-training of deep
  bidirectional transformers for language understanding,'' \emph{arXiv preprint
  arXiv:1810.04805}, 2018.

\bibitem{sukhbaatar2015end}
S.~Sukhbaatar, A.~Szlam, J.~Weston, and R.~Fergus, ``End-to-end memory
  networks,'' \emph{arXiv preprint arXiv:1503.08895}, 2015.

\bibitem{rocktaschel2015reasoning}
T.~Rockt{\"a}schel, E.~Grefenstette, K.~M. Hermann, T.~Ko{\v{c}}isk{\`y}, and
  P.~Blunsom, ``Reasoning about entailment with neural attention,'' \emph{arXiv
  preprint arXiv:1509.06664}, 2015.

\bibitem{sanh2019distilbert}
V.~Sanh, L.~Debut, J.~Chaumond, and T.~Wolf, ``Distilbert, a distilled version
  of bert: smaller, faster, cheaper and lighter,'' \emph{arXiv preprint
  arXiv:1910.01108}, 2019.

\bibitem{garvey2018framework}
C.~Garvey, ``A framework for evaluating barriers to the democratization of
  artificial intelligence,'' in \emph{Thirty-Second AAAI Conference on
  Artificial Intelligence}, 2018.

\bibitem{li2019edge}
E.~Li, L.~Zeng, Z.~Zhou, and X.~Chen, ``Edge ai: On-demand accelerating deep
  neural network inference via edge computing,'' \emph{IEEE Transactions on
  Wireless Communications}, vol.~19, no.~1, pp. 447--457, 2019.

\bibitem{xu2018scaling}
X.~Xu, Y.~Ding, S.~X. Hu, M.~Niemier, J.~Cong, Y.~Hu, and Y.~Shi, ``Scaling for
  edge inference of deep neural networks,'' \emph{Nature Electronics}, vol.~1,
  no.~4, pp. 216--222, 2018.

\bibitem{li2020ftrans}
B.~Li, S.~Pandey, H.~Fang, Y.~Lyv, J.~Li, J.~Chen, M.~Xie, L.~Wan, H.~Liu, and
  C.~Ding, ``Ftrans: energy-efficient acceleration of transformers using
  fpga,'' in \emph{Proceedings of the ACM/IEEE International Symposium on Low
  Power Electronics and Design}, 2020, pp. 175--180.

\bibitem{guo2020accelerating}
C.~Guo, B.~Y. Hsueh, J.~Leng, Y.~Qiu, Y.~Guan, Z.~Wang, X.~Jia, X.~Li, M.~Guo,
  and Y.~Zhu, ``Accelerating sparse dnn models without hardware-support via
  tile-wise sparsity,'' \emph{arXiv preprint arXiv:2008.13006}, 2020.

\bibitem{wang2020hat}
H.~Wang, Z.~Wu, Z.~Liu, H.~Cai, L.~Zhu, C.~Gan, and S.~Han, ``Hat:
  Hardware-aware transformers for efficient natural language processing,''
  \emph{arXiv preprint arXiv:2005.14187}, 2020.

\bibitem{ham20203}
T.~J. Ham, S.~J. Jung, S.~Kim, Y.~H. Oh, Y.~Park, Y.~Song, J.-H. Park, S.~Lee,
  K.~Park, J.~W. Lee \emph{et~al.}, ``A\^{} 3: Accelerating attention
  mechanisms in neural networks with approximation,'' in \emph{2020 IEEE
  International Symposium on High Performance Computer Architecture
  (HPCA)}.\hskip 1em plus 0.5em minus 0.4em\relax IEEE, 2020, pp. 328--341.

\bibitem{li2020_efficient}
B.~Li, Z.~Kong, T.~Zhang, J.~Li, Z.~Li, H.~Liu, and C.~Ding, ``Efficient
  transformer-based large scale language representations using
  hardware-friendly block structured pruning,'' in \emph{Proceedings of the
  2020 Conference on Empirical Methods in Natural Language Processing (EMNLP)},
  2020.

\bibitem{chen2020lottery}
T.~Chen, J.~Frankle, S.~Chang, S.~Liu, Y.~Zhang, Z.~Wang, and M.~Carbin, ``The
  lottery ticket hypothesis for pre-trained bert networks,'' \emph{arXiv
  preprint arXiv:2007.12223}, 2020.

\bibitem{prasanna-etal-2020-bert}
\BIBentryALTinterwordspacing
S.~Prasanna, A.~Rogers, and A.~Rumshisky, ``{W}hen {BERT} {P}lays the
  {L}ottery, {A}ll {T}ickets {A}re {W}inning,'' in \emph{Proceedings of the
  2020 Conference on Empirical Methods in Natural Language Processing
  (EMNLP)}.\hskip 1em plus 0.5em minus 0.4em\relax Online: Association for
  Computational Linguistics, Nov. 2020, pp. 3208--3229. [Online]. Available:
  \url{https://www.aclweb.org/anthology/2020.emnlp-main.259}
\BIBentrySTDinterwordspacing

\bibitem{chen2018best}
M.~X. Chen, O.~Firat, A.~Bapna, M.~Johnson, W.~Macherey, G.~Foster, L.~Jones,
  N.~Parmar, M.~Schuster, Z.~Chen \emph{et~al.}, ``The best of both worlds:
  Combining recent advances in neural machine translation,'' \emph{arXiv
  preprint arXiv:1804.09849}, 2018.

\bibitem{li2020efficient}
B.~Li, Z.~Kong, T.~Zhang, J.~Li, Z.~Li, H.~Liu, and C.~Ding, ``Efficient
  transformer-based large scale language representations using
  hardware-friendly block structured pruning,'' \emph{arXiv preprint
  arXiv:2009.08065}, 2020.

\bibitem{ma2020pconv}
X.~Ma, F.-M. Guo, W.~Niu, X.~Lin, J.~Tang, K.~Ma, B.~Ren, and Y.~Wang, ``Pconv:
  The missing but desirable sparsity in dnn weight pruning for real-time
  execution on mobile devices.'' in \emph{AAAI}, 2020, pp. 5117--5124.

\bibitem{2015CSR5}
W.~Liu and B.~Vinter, ``Csr5: An efficient storage format for cross-platform
  sparse matrix-vector multiplication,'' in \emph{The 29th ACM International
  Conference on Supercomputing (ICS '15)}, 2015.

\bibitem{1995Templates}
R.~E. Al., ``Templates for solution of linear systems: Building blocks for
  iterative methods,'' \emph{Siam}, 1995.

\bibitem{2009Pattern}
M.~Belgin, G.~·Ba·Ck, and C.~J. Ribbens, ``Pattern-based sparse matrix
  representation for memory-efficient smvm kernels,'' in \emph{International
  Conference on Supercomputing}, 2009, p. 100.

\bibitem{shi2020csb}
R.~Shi, P.~Dong, T.~Geng, Y.~Ding, X.~Ma, H.~K.-H. So, M.~Herbordt, A.~Li, and
  Y.~Wang, ``Csb-rnn: a faster-than-realtime rnn acceleration framework with
  compressed structured blocks,'' in \emph{Proceedings of the 34th ACM
  International Conference on Supercomputing}, 2020, pp. 1--12.

\bibitem{cao2019efficient}
S.~Cao, C.~Zhang, Z.~Yao, W.~Xiao, L.~Nie, D.~Zhan, Y.~Liu, M.~Wu, and
  L.~Zhang, ``Efficient and effective sparse lstm on fpga with bank-balanced
  sparsity,'' in \emph{Proceedings of the 2019 ACM/SIGDA International
  Symposium on Field-Programmable Gate Arrays}, 2019, pp. 63--72.

\bibitem{pinar1999improving}
A.~Pinar and M.~T. Heath, ``Improving performance of sparse matrix-vector
  multiplication,'' in \emph{SC'99: Proceedings of the 1999 ACM/IEEE Conference
  on Supercomputing}.\hskip 1em plus 0.5em minus 0.4em\relax IEEE, 1999, pp.
  30--30.

\bibitem{zachariadis2020accelerating}
O.~Zachariadis, N.~Satpute, J.~G{\'o}mez-Luna, and J.~Olivares, ``Accelerating
  sparse matrix--matrix multiplication with gpu tensor cores,'' \emph{Computers
  \& Electrical Engineering}, vol.~88, p. 106848, 2020.

\bibitem{kannan2013efficient}
R.~Kannan, ``Efficient sparse matrix multiple-vector multiplication using a
  bitmapped format,'' in \emph{20th Annual International Conference on High
  Performance Computing}.\hskip 1em plus 0.5em minus 0.4em\relax IEEE, 2013,
  pp. 286--294.

\bibitem{jiang2019accuracy}
W.~Jiang, X.~Zhang, E.~H.-M. Sha, L.~Yang, Q.~Zhuge, Y.~Shi, and J.~Hu,
  ``Accuracy vs. efficiency: Achieving both through fpga-implementation aware
  neural architecture search,'' in \emph{Proceedings of the 56th Annual Design
  Automation Conference 2019}, 2019, pp. 1--6.

\bibitem{jiang2019achieving}
W.~Jiang, E.~H.-M. Sha, X.~Zhang, L.~Yang, Q.~Zhuge, Y.~Shi, and J.~Hu,
  ``Achieving super-linear speedup across multi-fpga for real-time dnn
  inference,'' \emph{ACM Transactions on Embedded Computing Systems (TECS)},
  vol.~18, no.~5s, pp. 1--23, 2019.

\bibitem{jiang2019xfer}
W.~Jiang, X.~Zhang, E.~H.-M. Sha, Q.~Zhuge, L.~Yang, Y.~Shi, and J.~Hu, ``Xfer:
  A novel design to achieve super-linear performance on multiple fpgas for
  real-time ai,'' in \emph{Proceedings of the 2019 ACM/SIGDA International
  Symposium on Field-Programmable Gate Arrays}, 2019, pp. 305--305.

\bibitem{2018OuterSPACE}
S.~Pal, J.~Beaumont, D.~H. Park, A.~Amarnath, and R.~Dreslinski, ``Outerspace:
  An outer product based sparse matrix multiplication accelerator,'' in
  \emph{2018 IEEE International Symposium on High Performance Computer
  Architecture (HPCA)}, 2018.

\bibitem{zhao2019performance}
J.~Zhao, L.~Feng, S.~Sinha, W.~Zhang, Y.~Liang, and B.~He, ``Performance
  modeling and directives optimization for high level synthesis on fpga,''
  \emph{IEEE Transactions on Computer-Aided Design of Integrated Circuits and
  Systems}, 2019.

\bibitem{TP-toolbox-web}
Xilinx, ``Introduction to fpga design with vivado high-level synthesis,''
  \url{https://www.xilinx.com/support/documentation/sw_manuals/ug998-vivado-intro-fpga-design-hls.pdf}.

\bibitem{zoph2016neural}
B.~Zoph and Q.~V. Le, ``Neural architecture search with reinforcement
  learning,'' \emph{arXiv preprint arXiv:1611.01578}, 2016.

\bibitem{williams1992simple}
R.~J. Williams, ``Simple statistical gradient-following algorithms for
  connectionist reinforcement learning,'' \emph{Machine learning}, vol.~8, no.
  3-4, pp. 229--256, 1992.

\bibitem{yang2020co}
L.~Yang, Z.~Yan, M.~Li, H.~Kwon, L.~Lai, T.~Krishna, V.~Chandra, W.~Jiang, and
  Y.~Shi, ``Co-exploration of neural architectures and heterogeneous asic
  accelerator designs targeting multiple tasks,'' in \emph{2020 57th ACM/IEEE
  Design Automation Conference (DAC)}.\hskip 1em plus 0.5em minus 0.4em\relax
  IEEE, 2020, pp. 1--6.

\bibitem{merity2016pointer}
S.~Merity, C.~Xiong, J.~Bradbury, and R.~Socher, ``Pointer sentinel mixture
  models,'' \emph{arXiv preprint arXiv:1609.07843}, 2016.

\bibitem{wang2018glue}
A.~Wang, A.~Singh, J.~Michael, F.~Hill, O.~Levy, and S.~R. Bowman, ``Glue: A
  multi-task benchmark and analysis platform for natural language
  understanding,'' \emph{arXiv preprint arXiv:1804.07461}, 2018.

\bibitem{narang2017block}
S.~Narang, E.~Undersander, and G.~Diamos, ``Block-sparse recurrent neural
  networks,'' \emph{arXiv preprint arXiv:1711.02782}, 2017.

\bibitem{so2019evolved}
D.~So, Q.~Le, and C.~Liang, ``The evolved transformer,'' in \emph{International
  Conference on Machine Learning}.\hskip 1em plus 0.5em minus 0.4em\relax PMLR,
  2019, pp. 5877--5886.

\end{thebibliography}

\end{document}